\documentclass[preprint,review,10pt]{elsarticle}  %
\usepackage{amssymb}
\usepackage{bm}
\usepackage{amsmath}
\usepackage{indentfirst}
\usepackage{color}
\usepackage{epsfig}
\usepackage{booktabs}
\usepackage{algorithm}
\usepackage{algorithmicx}
\usepackage{algpseudocode}
\usepackage[separate-uncertainty = true,multi-part-units = repeat]{siunitx}

\usepackage[numbers]{natbib}
\usepackage{makecell}
\usepackage{multirow}
\usepackage{graphicx}
\usepackage{caption}
\usepackage{hyperref}
\usepackage[separate-uncertainty = true,multi-part-units = repeat]{siunitx}
\usepackage{xcolor}
\usepackage{mathtools}

\usepackage{lineno}

\begin{document}
	
	\begin{frontmatter}
	\title{Semi-Supervised Semantic Segmentation with Cross Teacher Training}

	\author[1]{Hui Xiao}
	\ead{2011082337@nbu.edu.cn}
	\cortext[cor1]{Corresponding author}
	\author[1]{Li Dong}
	\author[2]{Kangkang Song}
	\author[1]{Hao Xu}
	\author[3]{Shuibo Fu}
	\author[1]{Diqun Yan}
	\author[1]{Chengbin Peng\corref{cor1}}
	\ead{pengchengbin@nbu.edu.cn}
	
	\address[1]{Faculty of Electrical Engineering and Computer Science, Ningbo University, Ningbo, China}
	\address[2]{Ningbo Institute of Materials Technology \& Engineering, Chinese Academy of Sciences, Ningbo, China}
	\address[3]{Jiaochuan Academy, Ningbo, China}

	\begin{abstract}
		Convolutional neural networks can achieve remarkable performance in semantic segmentation tasks. However, such neural network approaches heavily rely on costly pixel-level annotation. Semi-supervised learning is a promising resolution to tackle this issue, but its performance still far falls behind the fully supervised counterpart. This work proposes a cross-teacher training framework with three modules that significantly improves traditional semi-supervised learning approaches. The core is a cross-teacher module, which could simultaneously reduce the coupling among peer networks and the error accumulation between teacher and student networks. In addition, we propose two complementary contrastive learning modules. The high-level module can transfer high-quality knowledge from labeled data to unlabeled ones and promote separation between classes in feature space. The low-level module can encourage low-quality features learning from the high-quality features among peer networks. In experiments, the cross-teacher module significantly improves the performance of traditional student-teacher approaches, and our framework outperforms state-of-the-art methods on benchmark datasets. Our source code of CTT will be released.
	\end{abstract}

	\begin{keyword}
		semantic segmentation \sep semi-supervised learning \sep student-teacher networks
	\end{keyword}

	\end{frontmatter}
	
	\section{Introduction}\label{}
	Semantic segmentation is a fundamental task in computer vision, which aims to assign a label to each pixel of an image. The development of deep convolutional neural networks has led to tremendous developments of semantic segmentation in many public benchmarks \cite{fu2019dual,long2015fully,li2020spatial,tao2020hierarchical,chen2017deeplab}. However, supervised semantic segmentation requires a large amount of pixel-level annotated data, but pixel-level annotations are expensive and time-consuming, which is a bottleneck in semantic segmentation.
	
\begin{figure}[htbp]

		\begin{center}
			
			\includegraphics[width=0.5\columnwidth]{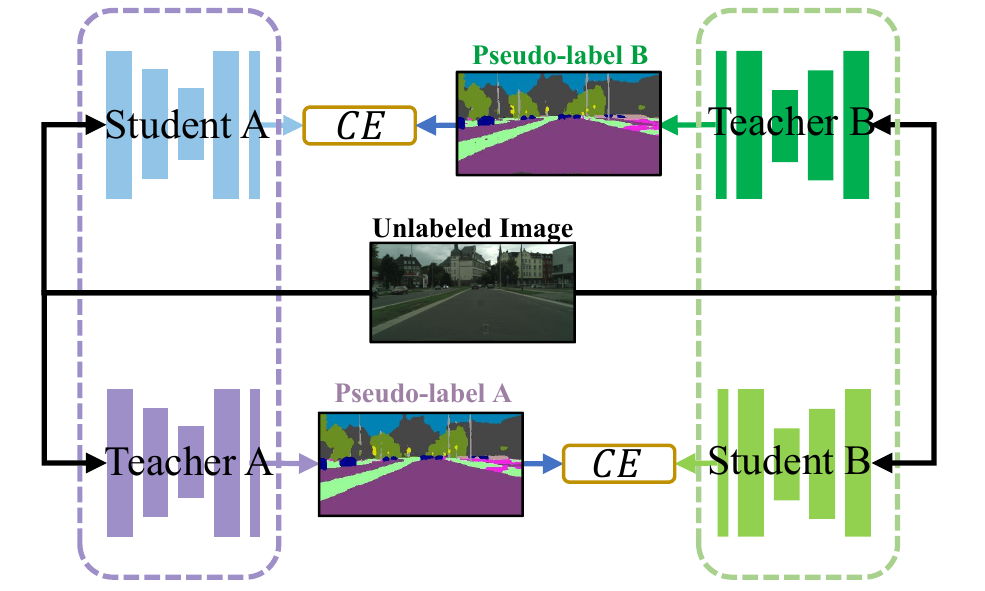}
			
		\end{center}
		
		\caption{The proposed Cross-Teacher Training (CTT) framework. Each dotted box indicates a pair of teacher-student networks.
			$CE$ is cross-entropy loss. The general idea is, in each iteration, a teacher network takes the exponential moving average of the student network in the same pair, and generates pseudo labels to teach the student network in another pair.  Multiple pairs of student-teacher networks are also allowed.}
		
		\label{cross learning}
		
	\end{figure}

	{
		Semi-supervised learning can alleviate this issue by extracting useful knowledge not only from labeled samples but also from unlabeled samples. Some semi-supervised semantic segmentation techniques generate pseudo labels for unlabeled samples and use entropy minimization to optimize the generation process  \cite{zhu2020improving,wei2021crest,teh2021gist,ke2020three,mittal2019semi,hung2018adversarial}.  Based on student-teacher models, some other approaches attempt to produce consistent labels from unlabeled samples with various data augmentation methods  to enhance consistency \cite{french2019semi,olsson2021classmix} or to filter out different types of noises with different peer networks to avoid over-fitting \cite{han2018co}.
		
		However, there are some limitations in existing semi-supervised learning approaches. For example, entropy minimization may make the network over-fit to erroneous pseudo labels  
		\cite{arazo2020pseudo}, and traditional student-teacher models usually suffer from network coupling issues \cite{ke2019dual}.
		In this work, we propose a Cross-Teacher Training (CTT) framework to address these problems for semantic segmentation. 
		It can reduce coupling between peer networks and reduce prediction bias accumulation. 
		As shown in Fig. \ref{cross learning}, pseudo labels for training each student network are generated by teacher networks from the other student-teacher networks.

		In addition, in semantic segmentation, segmentation networks often produce fuzzy contours and mislabel for rare objects in samples, as cross-entropy loss cannot promote intra-class compactness and enlarge inter-class boundaries. Furthermore, cross-entropy causes the performance of the network to significantly degrade as the number of labeled samples decreases \cite{zhao2020contrastive}.
		These problems are particularly prominent in semi-supervised learning because there are problems with little labeled data and pseudo-labels with noise.
		To address these problems, our CTT framework adds complementary contrastive learning modules to transfers feature-level knowledge from labeled data to unlabeled
		data and to encourage learning high-quality features from
		peer networks. }

	The main contributions can be summarized as follows:
	\begin{itemize} 
		\item We propose a cross-teacher training framework with
		a cross-teacher module that reduces erroneous knowledge accumulation in the learning process and reduces coupling between peer networks.
		
		\item We propose a high-level contrastive learning module
		that transfers knowledge from labeled to unlabeled
		data and improves inter-class separations in the feature
		space. In addition, we design a low-level contrastive
		learning module, which enables learning high-quality
		features among peer networks.
		\item  We demonstrates that, the cross-teacher module significantly improves the performance of traditional studentteacher approaches, and the whole framework outperforms state-of-the-art methods on benchmark datasets.
	\end{itemize}

	\section{Related Work}\label{}
	\subsection{Semantic Segmentation}
	The purpose of semantic segmentation is to identify objects in images at pixel level. 
	The proposal of full convolution neural network (FCN) \cite{long2015fully} is a milestone in semantic segmentation, which provides the basis for the follow-up development of semantic segmentation. {Recent methods have achieved excellent results by using dilated convolution \cite{chen2014semantic,chen2017deeplab,chen2017rethinking,chen2018encoder,yu2015multi}, deformable convolution \cite{dai2017deformable}, attention mechanism \cite{fu2019dual,huang2019ccnet,zhao2018psanet}, feature pyramid spatial pooling \cite{liu2015parsenet,zhao2018icnet,DBLP:conf/cvpr/LiuCTC21}, and encoder-decoder models \cite{badrinarayanan2017segnet,chen2018encoder,ronneberger2015u}.} {In addition, recently, semantic segmentation methods based on human hierarchy \cite{zhou2022Hierarchical} or prototypes \cite{zhou2022rethinking} have also received extensive attention. }Nevertheless, training these models usually requires lots of pixel-level annotations, and the acquisition of pixel-level annotation data is often expensive and time-consuming. These issues limit their practical application.
	
	\subsection{Semi-Supervised Semantic Segmentation}
	Pixel-level annotation is expensive. Semi-supervised approaches that exploit unlabeled data efficiently can alleviate this issue. Several studies have explored semi-supervised learning in semantic segmentation and have revealed promising results. S4GAN \cite{mittal2019semi} and AdvSemiSeg \cite{hung2018adversarial} select reliable predictions as pseudo labels by adding a discriminator network. ECS \cite{mendel2020semi} adopts the idea of self-correction, which enables the discriminator not only to learn to distinguish between correct and incorrect predictions but also to rectify current mistakes. GCT \cite{ke2020guided} combines mutual learning with error-correcting discriminators and uses different initialization weights or structures for each network to avoid network coupling yielding superior results. In addition, CCT \cite{ouali2020semi} performs consistency loss between a main decoder and auxiliary encoders with different perturbations, making decision boundaries be in low-density regions, and CPS \cite{chen2021semi} performs consistency loss between two networks. Based on mean teacher methods \cite{tarvainen2017mean}, Cutmix \cite{french2019semi} and Classmix \cite{olsson2021classmix} explore data augmentation methods suitable for semantic segmentation to enhance consistency. 
	DMT \cite{feng2020dmt} employs two models and utilizes inter-model inconsistencies to re-weight the loss and trains two models from different subsets of the label subset using difference maximization sampling.
	{SCF \cite{DBLP:conf/cvpr/IbrahimVRM20} uses an ancillary model that generates initial segmentation labels for the weak set and a self-correction module that improves the generated labels during training using the increasingly accurate primary model. 
		DCC \cite{DBLP:conf/cvpr/LaiTJ0ZWJ21} proposes to maintain the context-aware consistency for features generated from the same pixels but affiliated to different image patches. SemiSeg \cite{alonso2021semi}, $\text{C}^3$-SemiSeg \cite{DBLP:conf/iccv/0001X0GH21}, and P$\text{C}^2$Seg \cite{DBLP:conf/iccv/ZhongYWY0W21} explore the loss of consistency in the feature space. PseudoSeg \cite{zou2020pseudoseg} present a pseudo-labeling method to generate well-calibrated structured pseudo labels. }
	
	\subsection{Contrastive Learning}
	Contrastive learning is commonly used for learning similarity functions that bring augmented views of the same data closer and separate that views of different data farther  in the representation space. 
	In self-supervised learning, 
	Xie \emph{et al.} \cite{xie2021detco} applies hierarchical contrastive losses between whole images and local patches to improve object detection. 
	In supervised learning, 
	contrastive loss can be applied for outputs from different network layers \cite{bae2021self} or between image patches and their local contexts \cite{tarasiou2021context}.  Chen \emph{et al.} \cite{chen2021scnet} uses contrastive learning to associate pixels in a region with a prototype of that region. 
	Wang \emph{et al.} \cite{wang2021exploring} uses a set of pixels as memory regions to explore the pixel-to-area relationship. 
	{Zhou \emph{et al.} \cite{zhou2022regional} explores the contrastive learning algorithm used to solve semantic segmentation. Sel-CL \cite{Li2022SelCL} proposes selective-supervised contrastive learning for solving learning with noisy labels. CCSSL \cite{yang2022class} uses contrastive learning to improve the
		pseudo-label quality and enhance the robustness of models in real-world setting.} 
	Different from previous work, our contrastive learning not only facilitates class separation but also facilitates knowledge transfer from labeled data to unlabeled data and encourages learning high-quality features from peer networks.

	\begin{figure*}[t]
		
		\begin{center}
			
			\includegraphics[width=0.96\textwidth]{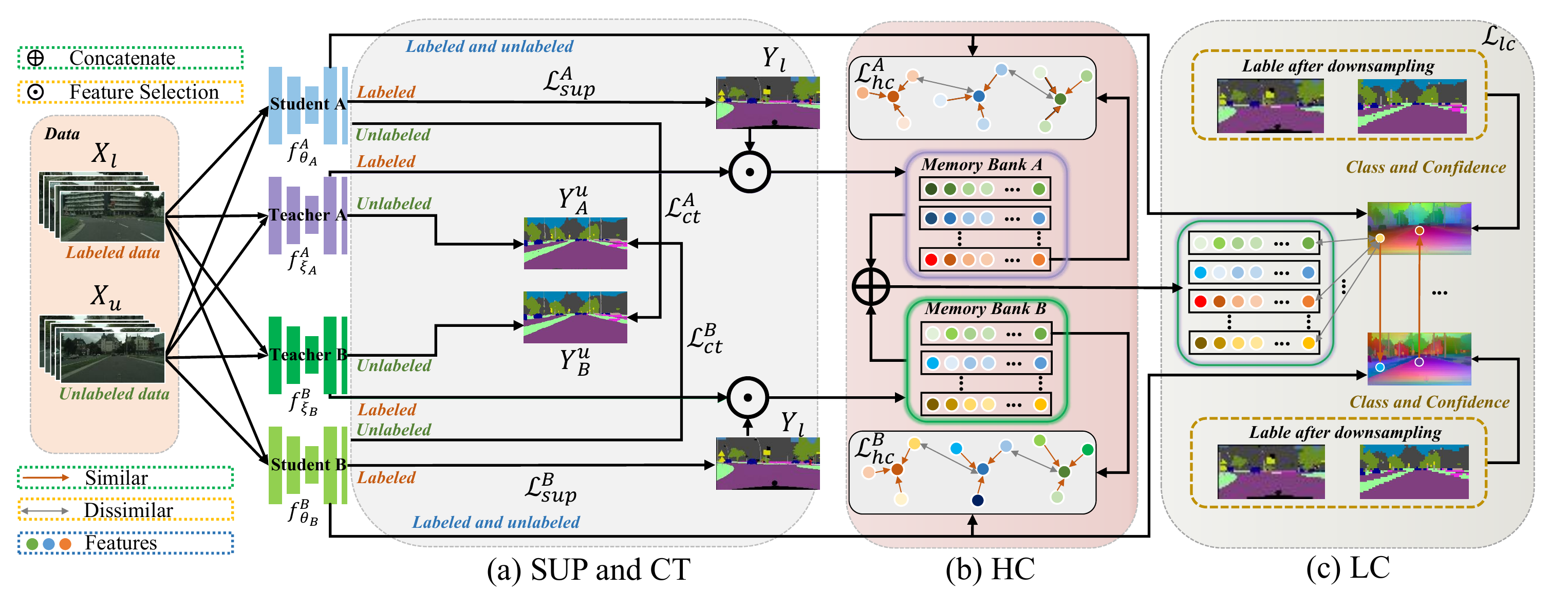}
			
		\end{center}
		
		\caption{Overview of our Cross Teacher Training (CTT) framework for semi-supervised semantic segmentation. 
			It contains two peer networks $\{f_{\theta_A}^{A},f_{\theta_B}^{B}\}$ and their corresponding teacher models $\{f_{\xi_A}^{A},f_{\xi_B}^{B}\}$.
			It consists of three parts, namely, (a) a Supervised Learning (SUP) module and a Cross-Teacher (CT) module, (b) a High-Level Contrastive Learning (HC) module, and (c) a Low-Level Contrastive Learning (LC) module. 
			In Part (a), given the unlabeled data, teacher models $\{f_{\xi_A}^{A},f_{\xi_B}^{B}\}$ generate the corresponding pseudo labels $\{Y_{A}^{u},Y_{B}^{u}\}$, and student models $\{f_{\theta_A}^{A},f_{\theta_B}^{B}\}$ are optimized by minimizing the unsupervised loss $\mathcal{L}_{ct}$ with the pseudo labels generated by the teacher of the peer model. The data follow represented by arrows is as follows: $f_{\xi_A}^{A}(X_u)\rightarrow{Y_{A}^{u}}\rightarrow{\mathcal{L}_{ct}}\leftarrow{f_{\theta_B}^{B}(X_u)}$ and $f_{\xi_B}^{B}(X_u)\rightarrow{Y_{B}^{u}}\rightarrow{\mathcal{L}_{ct}}\leftarrow{f_{\theta_A}^{A}(X_u)}$. 
			In Part (b), given the labeled data, teacher models select high-quality features to store in memory banks, and given the labeled and unlabeled data, student models generate the pixel-level features. We use two HC losses $\mathcal{L}_{hc}^A$ and $\mathcal{L}_{hc}^B$ to contrast pixel-level features with two memory banks respectively. 
			In Part (c), we use prediction confidence as a criterion for feature quality. We compare the confidence of the pixel-level features of the two student networks and use features with high confidence as positive keys. We use LC loss $\mathcal{L}_{lc}$ to optimize the network generating low-quality features to produce high-quality features.
		}
		
		\label{framework}
		
	\end{figure*}

	\section{Cross-Teacher Training Framework}
	
	In this work, we propose a new approach for semisupervised semantic segmentation called Cross-Teacher Training (CTT) framework. It includes a cross-teacher module that enables networks to adaptively correct training errors
	and learn decision boundaries from peer networks while reducing coupling between peer networks. It can have multiple
	pairs of student-teacher networks, and without loss of generality, we use two pairs in the following description. It also
	includes two complementary contrastive learning module to
	facilitate class separation and to promote feature quality. The
	overall model is shown in Fig. \ref{framework}.

	In semi-supervised semantic segmentation, it is assumed that there are lots of unlabeled data and a small amount of labeled data. Formally, the training dataset  $\mathcal{D}$ can be divided into two subsets: (1)  Labeled subset $\mathcal{D}_l=\{(X_{l}^{i},Y_{l}^{i})\}_{i=1}^{N_L}$ where $X_{l}^{i}$ is the $i$-th image, $Y_{l}^{i}$ represents the ground truth label corresponding to the images $X_{l}^{i}$, and $N_L$ is the number of labeled images; (2) Unlabeled subset $\mathcal{D}_u=\{X_{u}^{i}\}_{i=1}^{N_U}$, where $N_U$ is the number of unlabeled images. We consider that only a small fraction of the data is labeled, that is, $N_L\ll{N_U}$. With this setting, we develop a way to use $\mathcal{D}_l$ and  $\mathcal{D}_u$ efficiently to obtain a high-performance segmentation model.

	\subsection{Supervised Learning}
	First, we apply the cross-entropy to the labeled data $X_l$, as shown in Fig. \ref{framework}(a). Let $\mathcal{H}$ be the pixel-level cross-entropy loss function:
	\begin{align}
		{\hspace{-7pt}}			\mathcal{H}(P_s,Y_t)=-\frac{1}{H\times{W}}\sum_{h,w}\sum_{c}Y_{t}^{(h,w,c)}\log{(P_{s}^{(h,w,c)})}\text{,}
	\end{align}
	where $H\times{W}$ denotes the number of pixels, 
	$Y_t$ represents the one-hot vector transformed from the ground truth, and $P_s$ represents the prediction vector. Supervised loss for $f_{\theta_A}^A$ is defined as follows:
	\begin{equation}
		\begin{aligned}
			\mathcal{L}_{sup}^A=\mathcal{H}(f_{\theta_A}^A(X_l),Y_l)\text{,}
		\end{aligned}
	\end{equation}
	where $f_{\theta_A}^A$ denotes the student network and $\theta_A$ denotes the parameters of the student network. The total supervised loss function can be written as: $\mathcal{L}_{sup}=\mathcal{L}_{sup}^A+\mathcal{L}_{sup}^B$.

	\subsection{Cross-Teacher Module}
	To reduce erroneous knowledge accumulation in the learning process, we use a Cross-Teacher (CT) module in parallel with supervised learning, as shown in Fig. \ref{framework}(a).

	The method consists of two parallel segmentation networks and their corresponding teacher networks:
	\begin{align}
		P_{A}^u=f_{\theta_A}^{A}(X_{u}),Y_{A}^{u}= \text{one-hot}(f_{\xi_A}^{A}(X_{u}))\text{,}\\
		\label{PABYAB}
		P_{B}^u=f_{\theta_B}^{B}(X_{u}),Y_{B}^{u}= \text{one-hot}(f_{\xi_B}^{B}(X_{u}))\text{,}
	\end{align}
	where $P_{A}^u$ and $P_{B}^u$ are outputs of two student networks after softmax operation respectively.  These two networks have the same structure but with different initialization weights.  $Y_{A}^{u}$ and $Y_{B}^{u}$ are one-hot vectors representing class labels computed by two teacher networks respectively.
	
	We define $f_{\xi}$ as the teacher model obtained from exponential moving average (EMA) of the student model $f_{\theta}$, where $\xi$ is the EMA of the student network weights $\theta$ with a decay rate $\tau \in [0,1]$. This teacher model can be considered as a temporal ensemble of student models across training time $t$, thus producing more stable predictions for unlabeled images. At each training step, the weights of the teacher network $f_{\xi}$ are updated as follows: $\xi=\tau{\xi}+(1-\tau)\theta\text{.}$

	For unlabeled data $X_u$, we use  $Y_{A}^{u}$ computed by Eq. (3) as pseudo-labels, and to optimize network $f_{\theta_B}^{B}$, we minimize the cross-entropy loss between $Y_{A}^{u}$ and $P_B^u$. Similarly, optimizing network $f_{\theta_A}^{A }$ is to minimize the loss between  $Y_{B}^{u}$ and $P_A^u$. Note that the cross-entropy loss is not computed between peer networks $f_{\theta_A}^{A}$ and $f_{\theta_B}^{B}$ or within each pair of student-teacher networks, $f_{\theta_A}^{A }$-$f_{\xi_A}^{A }$ or $f_{\theta_B}^{B }$-$f_{\xi_B}^{B}$. This reduces coupling between peer networks and avoids error accumulation between teacher and student networks.

	The corresponding loss function is as follows:
	\begin{equation}
		\begin{aligned}
			\mathcal{L}_{ct}=&\mathcal{H}(P_{A}^u,Y_{B}^{u})
			+\mathcal{H}(P_{B}^u,Y_{A}^{u})\text{.}
		\end{aligned}
	\end{equation}

	\subsection{Complementary Contrastive Learning}
	Cross-entropy loss does not explicitly promote intra-class compactness or more significant boundaries between classes. 
	This factor can seriously affect the performance of segmentation. To address this problem, we propose two complementary contrastive learning modules with memory bank.

	\textbf{Memory Bank.} Memory banks store high-quality features from labeled data. 
	We expect the memory bank has representative features of all labeled data rather than the last few batches of data. Therefore, in our proposed memory bank, a first-in-first-out (FIFO) queue of size $N$ is created for each class. FIFO queue pushes $V$ features selected from the latest batch into the queue and pops out the outdated features at the top. Here, $V\ll{N}$. The definition of a memory bank $\mathcal{M}$ is as follows: $\mathcal{M}=\{Q_{c}\}_{c=1}^{|\mathcal{C}|}\text{,}$
	where $Q_{c}$ denotes the queue for class $c$, $\mathcal{C}$ is the set of categories, and  $|\mathcal{C}|$ denotes the number of classes. Therefore, our memory bank size is $|\mathcal{C}|\times{N}\times{D}$, where $N$ denotes the number of features as well as the FIFO queue size, and $D$ denotes feature dimension. Since there are two sets of networks in our framework, we need to maintain two different memory banks, $\mathcal{M}_A$ and $\mathcal{M}_B$.

	\textbf{ Feature Selection Criteria.}  
	To ensure the quality of the features in memory bank,
	our criteria for selecting features are as follows: (1) from the EMA model, (2) the features have the same label as the ground truth, and (3) top-$k$ features with the highest confidence in each batch. $k$ is the maximum number of features to be added to the memory bank in each iteration. During each iteration, for each class, candidate features are those that satisfy the Feature Selection Criteria. { We set $k=\max(1,N_L/N)$, where $N_L$ is number of labeled data. If the number of candidate features are less than $k$, then all the features are included in the memory bank. Otherwise, only $k$ randomly selected features are included.}

	\textbf{High-Level Contrastive Learning.} 
	To promote class separation and transfer knowledge from labeled to unlabeled data, we propose a High-Level Contrastive Learning (HC) module, as shown in Fig. \ref{framework}(b). 
	
	The general segmentation network $f=h\circ{g}$ consists of two parts: a feature extractor $h$ and a classifier $g$. In this network,  $h$ extracts pixel-level features  $\mathcal{Z}=h(X),\mathcal{Z}\in{\mathbb{R}^{h^{'}\times{w^{'}}\times{d}}}$ from image $X$, where $d$ is the length of vector. Classifier $g$ determines the classification probability of these features $\mathcal{G}=g(\mathcal{Z}), \mathcal{G}\in{\mathbb{R}^{h^{'}\times{w^{'}}\times{c}}}$, where $c$ is the length of vector. We define $\max(\text{softmax}(\mathcal{G}))$ as the confidence of features $\mathcal{Z}$.
	
	Formally, for features with high confidences, we use the HC module. Without loss of generality, for query feature  $i$ belonging to class $c$, where $i$ represents the feature corresponding to a pixel in a training image (including labeled data and unlabeled data) and $c$ comes from the pseudo label corresponding to the location of $i$, i.e., $c=\arg \max \text{softmax}(\mathcal{G}^{(i)})$ , this loss function encourages this feature to be similar to positive key $i^{+}$ in the class-$c$ queue and dissimilar to negative key $i^{-}$ in other queues. We define the HC loss function for $f_{\theta_A}^A$ as follows:
	\begin{align}
		\mathcal{L}_{hc}^A &= 
		-\frac{1}{\left|\mathcal{C}\right|}
		\sum_{c\in{\mathcal{C}}}\frac{1}{\left|\mathcal{Z}_c^A\right|}\sum_{i\in{\mathcal{Z}_c^A}}\frac{1}{\left|\mathcal{Q}_c^A\right|}\sum_{i^{+}\in{\mathcal{Z}'_c}}\nonumber\\
		&m_A^{(i)}\log\frac{\exp(i\cdot{i^{+}}/\tau)}
		{\exp(i\cdot{i^{+}}/\tau))+\sum \limits_{i^{-}\in{\mathcal{N}}}
			{\exp(i\cdot{i^{-}}/\tau)}}\text{,}
	\end{align}
	where 
	$\mathcal{Z}_{c}^A$ is a set of prediction features belonging to class $c$, $\mathcal{Q}_{c}^A$ is a set of features belonging to $c$-class queue in the memory bank $\mathcal{M}_A$, 
	$\tau$ is a temperature, 
	$\mathcal{N}$ denotes the features that do not belong to the $c$-class queue in the memory bank, and $m_A^{(i)}$ is a binary mask for feature $i$ defined as follows: 

	\begin{equation}
		\begin{aligned}
			m_A^{(i)}=\{\left\|f_{\theta_A}^A(X)\right\|_{\infty}^{(i)}>\phi\}_{\bm{1}}\text{,}
		\end{aligned}
	\end{equation}
	where $\{condition\}_{\bm{1}}$ returns $1$ when the $condition$ is true and $0$ otherwise, and $\left\|\cdot\right\|_{\infty}$ denotes the infinite norm. This definition indicates that we only select pixels with classification confidences higher than $\phi$ for contrastive training.  We denote that features with confidence higher than $\phi$ are high-quality, and those with confidence lower than $\phi$ are low-quality. Feature classifications are obtained from their corresponding ground truth labels or pseudo-labels. The $\mathcal{L}_{hc}^B$ loss of network $f_{\theta_B}^B$ is similar to that of $\mathcal{L}_{hc}^A$. The total HC loss function can be written as: $\mathcal{L}_{hc}=\mathcal{L}_{hc}^{A}+\mathcal{L}_{hc}^{B}$.

	\begin{algorithm}[!ht]
		\begin{minipage}{0.9\linewidth}
		\caption{The algorithm of our proposed CTT}
		\label{alg:algorithm}
		\textbf{Input}: Labeled dataset $\mathcal{D}_l$, unlabeled dataset $\mathcal{D}_u$, loss weight $\lambda_{sup}$, $\lambda_{ct}$, $\lambda_{hc}$, $\lambda_{lc}$, memory size $N$, confidence parameter $\phi$, max epochs $T_{max}$\\
		\textbf{Initialization}: Parameters of networks $\theta_A$(student A), $\xi_A$(teacher A), $\theta_B$(student B), $\xi_B$(teacher B), memory bank $\mathcal{M}_A$, $\mathcal{M}_B$\\
		\textbf{Output}: $\theta$
		\begin{algorithmic}[1] 
			\For{$t=1,2,\cdots,T_{max}$}
			\State Compute supervised loss $\mathcal{L}_{sup}$ by Eq. {2}
			\State Generate pseudo-labels $P_A^u$ and $P_B^u$ by Eq. {3} and {4}
			\State Compute cross-teacher loss $\mathcal{L}_{ct}$ by Eq. {5}
			\State Update memory banks by Feature Selection Criteria 
			\If{Memory Bank is filled up}
			\State Generate high-level loss mask and low-level
			\State loss mask by Eq. {7} and Eq. {9} respectively
			\State Compute high-level loss $\mathcal{L}_{hc}$ by Eq. {6}
			\State Compute low-level loss $\mathcal{L}_{lc}$ by Eq. {8}
			\State Total loss $\mathcal{L}_{total}\gets\lambda_{sup}\mathcal{L}_{sup}+\lambda_{ct}\mathcal{L}_{ct}
			+\lambda_{hc}\mathcal{L}_{hc}$
			\State $+\lambda_{lc}\mathcal{L}_{lc}$
			\Else
			\State Total loss $\mathcal{L}_{total}\gets\lambda_{sup}\mathcal{L}_{sup}+\lambda_{ct}\mathcal{L}_{ct}$
			\EndIf
			\State /* $\theta$ includes $\theta_A$ and $\theta_B$, $\xi$ includes $\xi_A$ and $\xi_B$ */
			\State Update $\theta(t+1)\gets\theta(t)+\alpha\nabla_{\theta}\mathcal{L}_{total}(\theta)$
			\State Update $\xi(t+1)\gets\tau\xi(t)+(1-\tau)\theta(t+1)$

			\EndFor
			
			\State \textbf{return} $\theta$

		\end{algorithmic}
		\end{minipage}
	\end{algorithm}

\begin{figure}[t]
		
		\begin{center}
			
			\includegraphics[width=0.5\columnwidth]{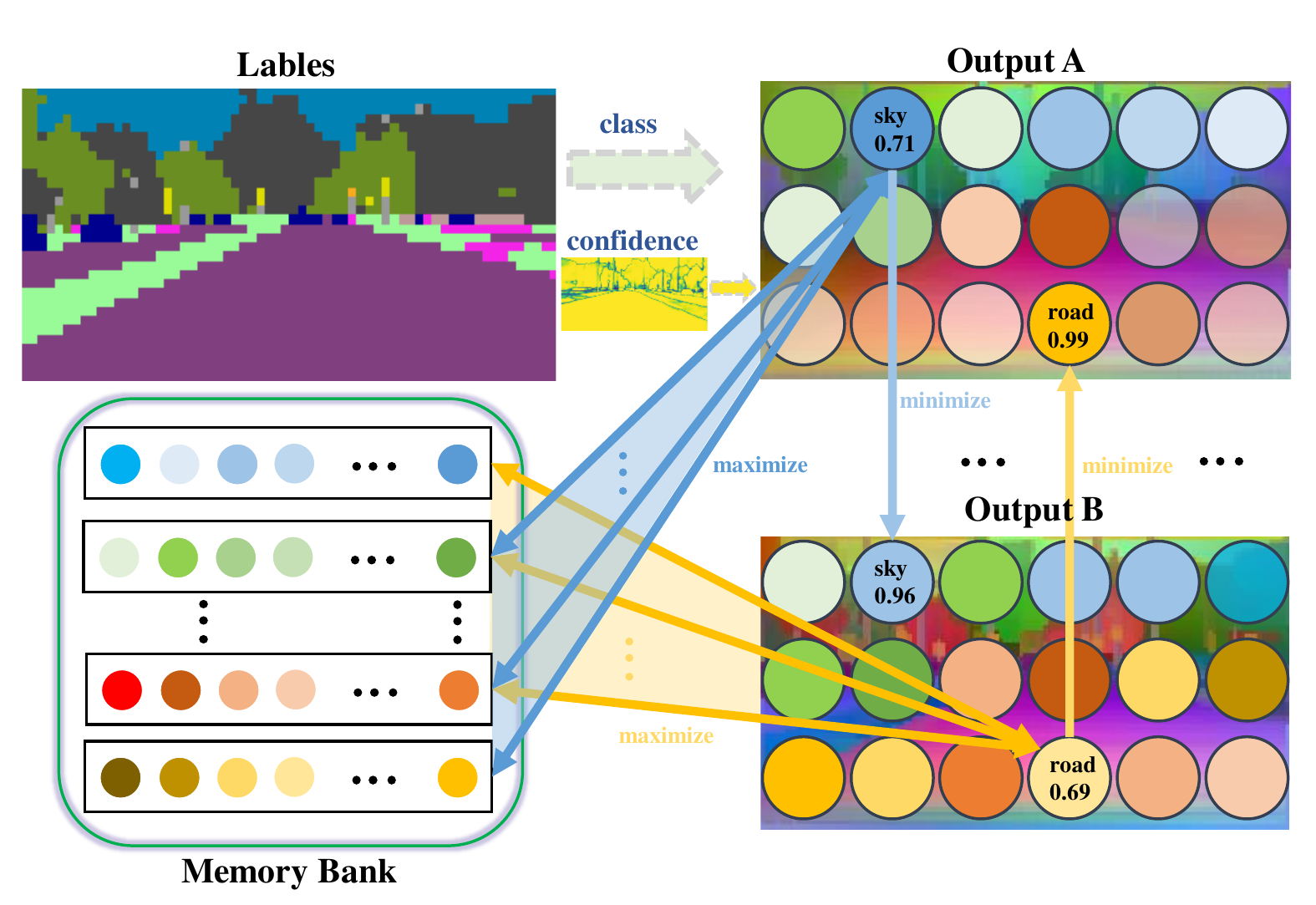}
			
		\end{center}
		
		\caption{Illustration of Low-Level Contrastive Learning. 
			The two subfigures on the right represent the outputs of features extracted from different networks, respectively. 
			We teach the network generating low-confidence features (LCF, e.g., 0.69 in the figure) by minimizing the distances between LCF and high-confidence features (HCF, e.g., 0.99 in the figure). We also teach networks to generate features far away from those with different classes in the memory bank.  
		}
		\label{cross}
		
	\end{figure}

	\textbf{Low-Level Contrastive Learning.} 
	To encourage low-quality features learning from high-quality features among peer networks, we propose a Low-Level Contrastive Learning (LC) module, as shown in Fig. \ref{framework}(c). For a same image, each network has a different region of interest.  
	Consequently, we can select the best quality features by comparing among peer networks. 
	Features with higher confidence are usually more accurate. 
	Therefore, guiding low confidence features by high confidence features can effectively facilitate the separation of classification boundaries. Based on the HC module, it is very convenient to introduce LC module, as shown in Fig. \ref{cross}. The LC loss $\mathcal{L}_{lc}$ for $f_{\theta_A}^A$ is defined as follows: 
	\begin{align}
		\mathcal{L}&_{lc}^{(A,B)} = 
		-\frac{1}{\left|\mathcal{C}\right|}
		\sum_{c\in{\mathcal{C}}}\frac{1}{\left|\mathcal{Z}_{c}^{A}\right|}\sum_{i\in{\mathcal{Z}_{c}^{A}}}\nonumber\\
		&m_{(A,B)}^{(i)}\log\frac{\exp(i\cdot{i^{+}}/\tau)}
		{\exp(i\cdot{i^{+}}/\tau)+\sum \limits_{i^{-}\in{\sum{\mathcal{N}}}}
			\exp(i\cdot{i^{-}}/\tau)}\text{,}
		\label{Lccdef}
	\end{align}
	where positive key $i^{+}$ is a feature obtained from network $f_{\theta_B}^{B}$ at the same pixel as feature $i$, negative key $i^{-}$ is selected from two memory banks corresponding to the two pairs of networks, and $m_{(A,B)}^{(i)}$ is a binary mask. Because $i^{+}$ is a contrast feature, we do not back-propagate loss through network $f_{\theta_B}^{B}$. 
	We define the pixel level mask $m_{(A,B)}^{(i)}$ as follows:
	
	\begin{align}
		{\hspace{-6pt}}			m_{(A,B)}^{(i)}=(1-{m}_A^{(i)}) \{\left\|f_{\theta_A}^{A}(X)\right\|_{\infty}^{(i)}<\left\|f_{\theta_B}^{B}(X)\right\|_{\infty}^{(i)}\}_{\bm{1}},
	\end{align}
	indicating that a pixel is taken into account only if the corresponding $m^{(i)}$ is zero and network $f_{\theta_A}^A$ has lower confidence than network $f_{\theta_B}^B$ on that pixel. Finally, the LC loss function can be written as: $\mathcal{L}_{lc}=\mathcal{L}_{lc}^{(A,B)}+\mathcal{L}_{lc}^{(B,A)}$.

	With this definition, the algorithm selects positive keys that can generate high confidence predictions to train networks that  generate low confidence predictions, forming a cross-learning scheme. It can also enlarge distances between each feature and its corresponding negative keys learned from both pairs of networks. The combination of negative keys from more than one network can increase the number of negative keys, enhancing the training process.
	
	\subsection{Training Process}
	Training is performed by optimizing the following loss function:
	\begin{align}
		\mathcal{L}_{total}=&\lambda_{sup}\mathcal{L}_{sup}+\lambda_{ct}\mathcal{L}_{ct}
		+\lambda_{hc}\mathcal{L}_{hc}+\lambda_{lc}\mathcal{L}_{lc}\text{,}
	\end{align}
	where $\lambda_{sup}$ denotes the weight of supervised loss, $\lambda_{ct}$ denotes the weight of CT loss, $\lambda_{hc}$ denotes the weight of HC loss and $\lambda_{lc}$ denotes the weight of LC loss. In experiment, $\mathcal{L}_{sup}$ and $\mathcal{L}_{ct}$ have been optimized simultaneously at the beginning of the training, but $\mathcal{L}_{hc}$ and $\mathcal{L}_{lc}$ have only been optimized when each class queue in the memory bank is filled. The overall algorithm process is detailed in Algorithm. \ref{alg:algorithm}.

	\section{Experimental Results}

	\begin{table*}[!htbp]
		
		\caption{Comparison with state-of-the-art methods on Cityscapes
			val set under different partition protocols. We randomly select 1/30 (100 images), 1/8 (372 images), 1/4 (744 images), and 1/2 (1487 images) of the training set in Cityscapes as labeled data, respectively, and the rest as unlabeled data. The performance gap with Full Supervised (FS) is shown in parentheses. All methods use DeepLabV2 and ResNet101.}
		\centering
		\label{city}
		\resizebox{12cm}{!}{
			\begin{tabular}{l|ccccc}  
				
				\toprule   
				\textbf{Methods} & 1/30 (100) & 1/8 (372) & 1/4 (744) & 1/2 (1487) &FS (2975)\\  
				\midrule
				\midrule   
				Sup. baseline (DeepLabV2+ResNet101) & 48.36 (-18.40) & 57.74 (-9.02) & 61.63 (-5.13) & 63.13 (-3.63) & 66.76\\
				AdvSemiSeg \cite{hung2018adversarial} & - & 58.80 (-7.60) & 62.30 (-4.10) & 65.70 (-0.70) & 66.40 \\
				s4GAN \cite{mittal2019semi} & - & 59.30 (-6.70) & 61.90 (-4.10) & - & 66.00 \\
				French \emph{et al.} \cite{french2019semi} & 51.20 (-16.33) & 60.34 (-7.19) & 63.87 (-3.66) & - & 67.53\\
				DST-CBC \cite{feng2020semi} & 48.70 (-18.20) & 60.50 (-6.40)& 64.40 (-2.50) & - & 66.90 \\
				ClassMix \cite{olsson2021classmix} & 54.07 (-12.12) & 61.35 (-4.84)& 63.63 (-2.56) & 66.29 (+0.10) & 66.19\\
				DMT \cite{feng2020dmt} & 54.80 (-13.36) & 63.03 (-5.31)& -  & - & 68.16\\
				ECS \cite{mendel2020semi} & - & 60.26 (-) & 63.77 (-) & - & - \\
				$\text{C}^3$-SemiSeg \cite{DBLP:conf/iccv/0001X0GH21} & 55.17 (-11.70) & 63.23 (-3.64) & 65.50 (-1.37) & - & 66.87 \\

				\midrule
				\textbf{CTT}  & \textbf{59.53 (-7.23)} & \textbf{64.04 (-2.72)} & \textbf{65.83 (-0.93)}  & \textbf{67.83 (+1.07)} & 66.76\\
				\bottomrule  
			\end{tabular}
		}
	\end{table*}
	
	In this section, we perform a series of experiments to validate the effectiveness and superiority of the proposed method.
	\subsection{Experiment Setup}
	\textbf{Datasets.} 
	We conduct experiments on benchmark datasets: Cityscapes \cite{cordts2016cityscapes} and PASCAL VOC 2012 \cite{everingham2015pascal}. Cityscapes contains 5000 images of real-world urban traffic scenes with image size $1024\times2048$ and each pixel belongs to one of nineteen classes. With this data set, we use 2975 images for training, 500 images for validation, and 1525 images for testing. PASCAL VOC 2012 is a natural landscape dataset with twenty-one classes including a background class. With this dataset, we use 10582 images for training and 1449 images for validation.

	\textbf{Backbone.} We use the widely used DeepLab \cite{chen2017deeplab} network in our experiments. We use DeepLabv2 or Deeplabv3+ with ResNet-101 \cite{he2016deep} as the backbone by default. 
	
	\textbf{Implementation Details.} We implement our method using PyTorch \cite{paszke2017automatic} and conduct experiments by following settings of the recent researches \cite{olsson2021classmix,hung2018adversarial,mittal2019semi}. We use the data augmentation methods in all the experiments, including rotations, flips, and translations. 
	
	All networks are trained with a stochastic gradient descent (SGD) optimizer. The initial learning rate is $2.5\times{e^{-4}}$, and the momentum is  $0.9$. A polynomial decay strategy, i.e., $1-(\frac{\text{iter}}{\text{max\_iter}})^{\text{power}}$ with power $0.9$, is used to adjust the learning rate. During training, we downsample each image to $512\times1024$ and randomly crop it to $256\times512$ with batch size 2 for the Cityscapes dataset and we use a crop size of $321\times321$ with batch size 10 for the PASCAL VOC 2012 dataset. {All experiments are trained for 80k iterations, which is 100 epochs for Cityscapes and 150 epochs for PASCAL VOC 2012, respectively.} The loss weights and hyper-parameters are as follows: $\lambda_{sup}=1$, $\lambda_{ct}=1$, $\lambda_{hc}=0.1$, $\lambda_{lc}=0.1$, $\alpha=0.99$, $\tau=0.5$, $N=128$, and $\phi=0.75$.
	In our experiments, the features with the highest confidence in each batch are selected to be added to the memory bank. To save computational costs, we perform complementary contrastive learning directly on the network output instead of the upsampled output, so we need to conduct downsampling for labels and pseudo-labels. As in previous methods, we use mean Intersection over Union (mIOU) to evaluate the performance of our method.

	\subsection{Results}
	\textbf{Results on Cityscapes Dataset.} We verified the effectiveness of our method in Cityscapes. For fairness, all comparison methods use DeepLabV2 and ResNet101. In each experiment, we randomly select 1/30 (100 images), 1/8 (372 images), 1/4 (744 images), or 1/2 (1487 images) of the training set in Cityscapes as labeled data and the remaining part of the training set as unlabeled data. Table \ref{city} shows the comparative results. 
	Our method outperforms existing methods with significant superiority. In columns 1/30, 1/8, 1/4 and 1/2, our method improves mIoU by 11.17\%, 6.30\%, 4.20\% and 4.70\%, respectively, comparing with baseline approaches using labeled data only. CTT exceeds performance of full supervision when partitioning 1/2 of the Cityscapes dataset into labeled data, exceeding 1.07\% mIou.
	
\begin{figure}[t]
		
		\begin{center}
			
			\includegraphics[width=0.9\columnwidth]{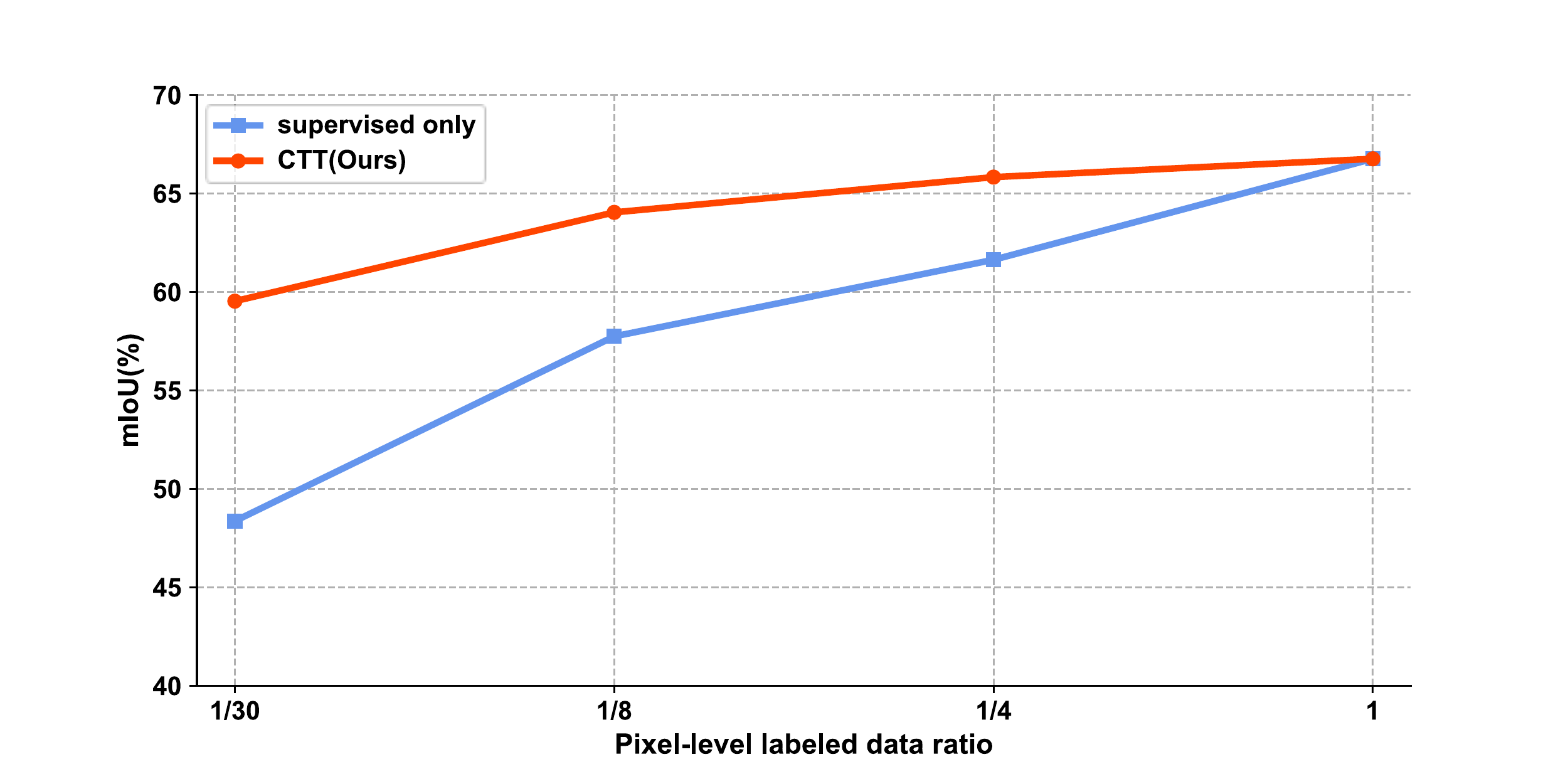}
			
		\end{center}
		
		\caption{
			Improvement over the supervised baseline, in a semi-supervised setting (w/
			unlabeled data) on Cityscapes val. All results are based on DeepLabV2 with ResNet101.
		}
		\label{sup_ctt}
		
	\end{figure}

\begin{figure}[t]
		
		\begin{center}
			
			\includegraphics[width=0.9\columnwidth]{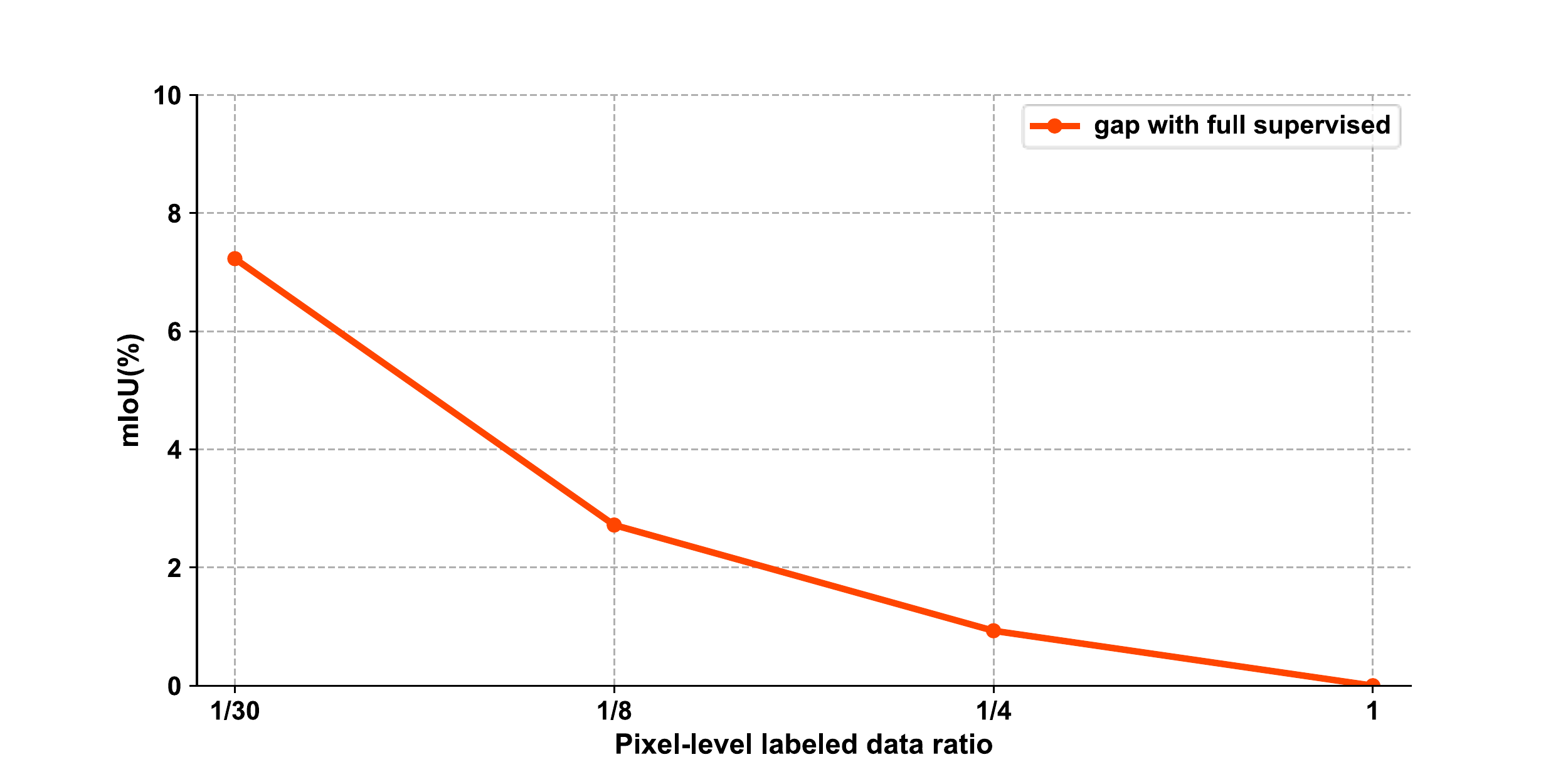}
			
		\end{center}
		
		\caption{
			Gap with full supervised, in a semi-supervised setting (w/
			unlabeled data) on Cityscapes val. All results are based on DeepLabV2 with ResNet101.
		}
		\label{ctt_gap}
		
	\end{figure}

	Fig. \ref{sup_ctt} presents the gap between CTT and baseline at different proportions of labeled data. 
	We observes that the less available labeled data, the higher gap between CTT and baseline.
	The performance of our method is also closest to that of the Full Supervised approaches which use the whole training set as labeled data. Fig. \ref{ctt_gap} shows the gap between CTT and fully supervised with different number of labeled samples. As the number of labeled samples increases, CTT gets closer to the performance of full supervised.

	{
		\textbf{Results on PASCAL VOC 2012 Dataset. } We further evaluate the effectiveness of our method on the PASCAL VOC 2012 dataset. For fairness, all comparison methods use DeepLabV2 and ResNet101. In each experiment, we randomly select 1/50 (212 images), 1/20 (529 images), 1/8 (1322 images), or 1/4 (2645 images) of the training set in PASCAL VOC 2012 as labeled data, and the remaining part of the training set as unlabeled data. Table \ref{voc} shows the comparison results with other methods. Table \ref{voc} reveals that CTT outperforms other methods in most cases. Even if CTT uses only a small portion of labeled data (e.g., 1/50), it still yields surprising results. This confirms the effectiveness of CTT in a semi-supervised setting. More notably, CTT exceeds the performance of full supervision when partitioning 1/4 of the PASCAL VOC 2012 dataset into labeled data.
		
		\textbf{Results on classic PASCAL VOC 2012 Dataset. } Table \ref{vocoff} compares our method with other state-of-the-art methods on the classic PASCAL VOC 2012 dataset. The experimental settings follows the PseudoSeg \cite{zou2020pseudoseg}. PseudoSeg randomly selects 1/2,1/4,1/8, and 1/16 of the classic VOC training set (1.4k images) as labeled data, and the remaining data and augmented set (9k images) as unlabeled data. For fairness, all comparison methods use DeepLabV3+ and ResNet101. Our method CTT outperforms PseudoSeg for all partitioning protocols, reaching +6.40\%, +5.55\%, +2.58\% and +3.22\% for 1/16, 1/8, 1/4 and 1/2 partitioning protocols, respectively.
		
	}
	
	\begin{table*}[!htbp]
		
		\caption{Comparison with state-of-the-art methods on PASCAL VOC 2012
			val set under different partition protocols. We randomly select 1/50 (212 images), 1/20 (529 images), 1/8 (1322 images), and 1/4 (2645 images) of the training set in PASCAL VOC 2012 as labeled data and the rest as unlabeled data. The performance gap with Full Supervised (FS) is shown in parentheses. All methods use DeepLabV2 and ResNet101.}
		\centering
		\label{voc}
		\resizebox{12cm}{!}{
			\begin{tabular}{l|ccccc}

				\toprule   
				\textbf{Methods} & 1/50 (212) & 1/20 (529) & 1/8 (1322) & 1/4 (2645) &FS (10582)\\  
				\midrule
				\midrule   
				Sup. baseline (DeepLabV2+ResNet101) & 55.69 (-18.57) & 61.36 (-12.90) & 67.14 (-7.12) & 70.20 (-4.06) & 74.26\\
				AdvSemiSeg \cite{hung2018adversarial} & 57.20 (-17.70) & 64.70 (-10.20) & 69.50 (-5.40) & 72.10 (-2.80) & 74.90 \\
				s4GAN \cite{mittal2019semi} & 63.30 (-10.30) & 67.20 (-6.40) & 71.40 (-2.20) & - & 66.00 \\
				French \emph{et al.} \cite{french2019semi} & 64.81 (-7.78) & 66.48 (-6.11) & 67.60 (-4.99) & - & 72.59\\
				DST-CBC \cite{feng2020semi} & 65.50 (-8.10) & 69.30 (-4.30)& 70.70 (-2.90) & 71.80 (-1.80) & 73.60 \\
				ClassMix \cite{olsson2021classmix} & 66.15 (-7.98) & 67.77 (-6.36)& 71.00 (-3.13) & 72.45 (-1.68) & 74.13\\
				DMT \cite{feng2020dmt} & 67.15 (-7.60) & 69.92 (-4.83)& 72.70 (-2.05)  & - & 74.75\\
				ECS \cite{mendel2020semi} & - & - & 70.22 (-6.07) & 72.60 (-3.69) & 76.29 \\
				
				GCT \cite{ke2020guided} & - & - & 72.14 (-3.18) & 73.62 (-1.70) & 75.32 \\
				SemiSeg-Contrastive \cite{alonso2021semi} & 67.90 (-6.20) & 70.0 (-4.10) & 71.6 (-2.50) & - & 74.10 \\
				\midrule
				\textbf{CTT}  & \textbf{69.24 (-5.02)} & \textbf{72.02 (-2.24)} & \textbf{73.66 (-0.60)}  & \textbf{75.07 (+0.81)} & 74.26\\
				\bottomrule  
			\end{tabular}
		}
	\end{table*}

	\begin{table}[!htbp]
		
		\setlength\tabcolsep{1pt}  
		\caption{Comparison with state-of-the-art methods on classic PASCAL VOC 2012 val set under different partition protocols. We follow the same partition protocols provided in PseudoSeg \cite{zou2020pseudoseg}. We randomly select 1/16 (92 images), 1/8 (183 images), 1/4 (366 images), and 1/2 (732 images) of the training set in classic PASCAL VOC2012 as labeled data, respectively, and the rest as unlabeled data. All methods use DeepLabV3+ and ResNet101.
		}
		\centering
		\label{vocoff}
		\scalebox{0.69}{
			\begin{tabular}{l|ccccc}  
				
				\toprule   
				\textbf{Methods} & 1/16 (92) & 1/8 (183) & 1/4 (366) & 1/2 (732) \\  
				\midrule
				\midrule   
				AdvSemSeg \cite{hung2018adversarial} & 39.69 & 47.58 & 59.97 & 65.27 \\
				CCT \cite{ouali2020semi} & 33.10 & 47.60 & 58.80 & 62.10 \\
				GCT \cite{ke2020guided} & 46.04 & 54.98 & 64.71 & 70.67 \\
				VAT \cite{DBLP:journals/pami/MiyatoMKI19} & 36.92 & 49.35 & 56.88 & 63.34 \\
				CutMix \cite{french2019semi} & 55.58 & 63.20 & 68.36 & 69.84 \\
				PseudoSeg \cite{zou2020pseudoseg} & 57.60 & 65.50 & 69.14 & 72.41 \\
				P$\text{C}^2$Seg \cite{DBLP:conf/iccv/ZhongYWY0W21} & 57.00 & 66.28 & 69.78 & 73.05 \\
				\midrule
				Sup. baseline & 41.22 & 50.82 & 60.37 & 66.29 \\
				\textbf{CTT}  & \textbf{64.00} & \textbf{71.05} & \textbf{72.36}  & \textbf{76.13} \\
				\bottomrule  
			\end{tabular}
		}
	\end{table}

	\subsection{Ablation Study}
	In this part, we investigate the performance gain of each proposed module through ablation experiments. All the following experiments are performed on the Cityscapes dataset with $1/8$ labeled data. All experiments use DeeplabV2 and ResNet101.
	
	\begin{table}[h]
		
		\setlength\tabcolsep{15pt}  
		\caption{Contribution of each module. $\mathcal{L}_{sup}$ indicates supervised loss. $\mathcal{L}_{ct}$ indicates unsupervised loss, i.e., cross-teacher loss. $\mathcal{L}_{hc}$ indicates high-level contrastive loss. $\mathcal{L}_{lc}$ indicates low-level contrastive loss. Results are obtained on the Cityscapes dataset with $1/8$ labeled data, DeepLabV2-ResNet101.}
		\centering
		\label{module}
		\scalebox{0.69}{
			\begin{tabular}{cccc|c}  
				\toprule   
				$\mathcal{L}_{sup}$ & $\mathcal{L}_{ct}$ & $\mathcal{L}_{hc}$ & $\mathcal{L}_{lc}$ & mIoU\%\\
				\midrule
				\midrule 
				\checkmark & &  &  & 57.74\\
				\checkmark & \checkmark &  &  & 63.69\\
				\checkmark & & \checkmark &  & 60.79\\
				\checkmark & & & \checkmark & 61.31\\
				\checkmark & & \checkmark & \checkmark & 62.30\\
				\checkmark & \checkmark & & \checkmark & 63.97\\
				\checkmark & \checkmark & \checkmark & & 63.50\\
				\midrule 
				\checkmark & \checkmark & \checkmark & \checkmark & \textbf{64.04}\\
				\bottomrule  
			\end{tabular}
		}
	\end{table}
	
	We first show the contribution of each proposed module to the semi-supervised performance in Table \ref{module}. As shown in Table \ref{module}, each module remarkably improves the performance of semi-supervised learning. Compared to the baseline $\mathcal{L}_{sup}$ trained on labeled data only, cross-teacher with $\mathcal{L}_{ct}$ individually yields 63.69\% in mIoU, bringing improvements of 6.7\%. Meanwhile, high-level contrastive learning $\mathcal{L}_{hc}$ and low-level contrastive learning $\mathcal{L}_{lc}$ outperform the baseline by 3.81\% and 4.35\%, respectively. When we integrate these modules, the performance is improved further to 64.04\%. The results show that cross-teacher, high-level contrastive learning, and low-level contrastive learning are beneficial for semi-supervised semantic segmentation.
	
	{\textbf{Effect of using n-pairs student-teacher network.} We investigate the impact of different pairs of teacher-student networks on performance, as shown in Table \ref{n-pairs}. We compare experiments on using one pair (traditional approach),  two pairs (our approach), and three pairs (extension of our  approach, { in which the student network in each pair is trained using pseudo labels generated by teacher networks in other pairs }) of teacher-student networks  in Table \ref{n-pairs}, and experimental results verifies the effectiveness of multiple pairs of  teacher-student networks.
	}
	
	\begin{figure*}[!htbp]
		
		\begin{center}
			
			\includegraphics[width=0.96\textwidth]{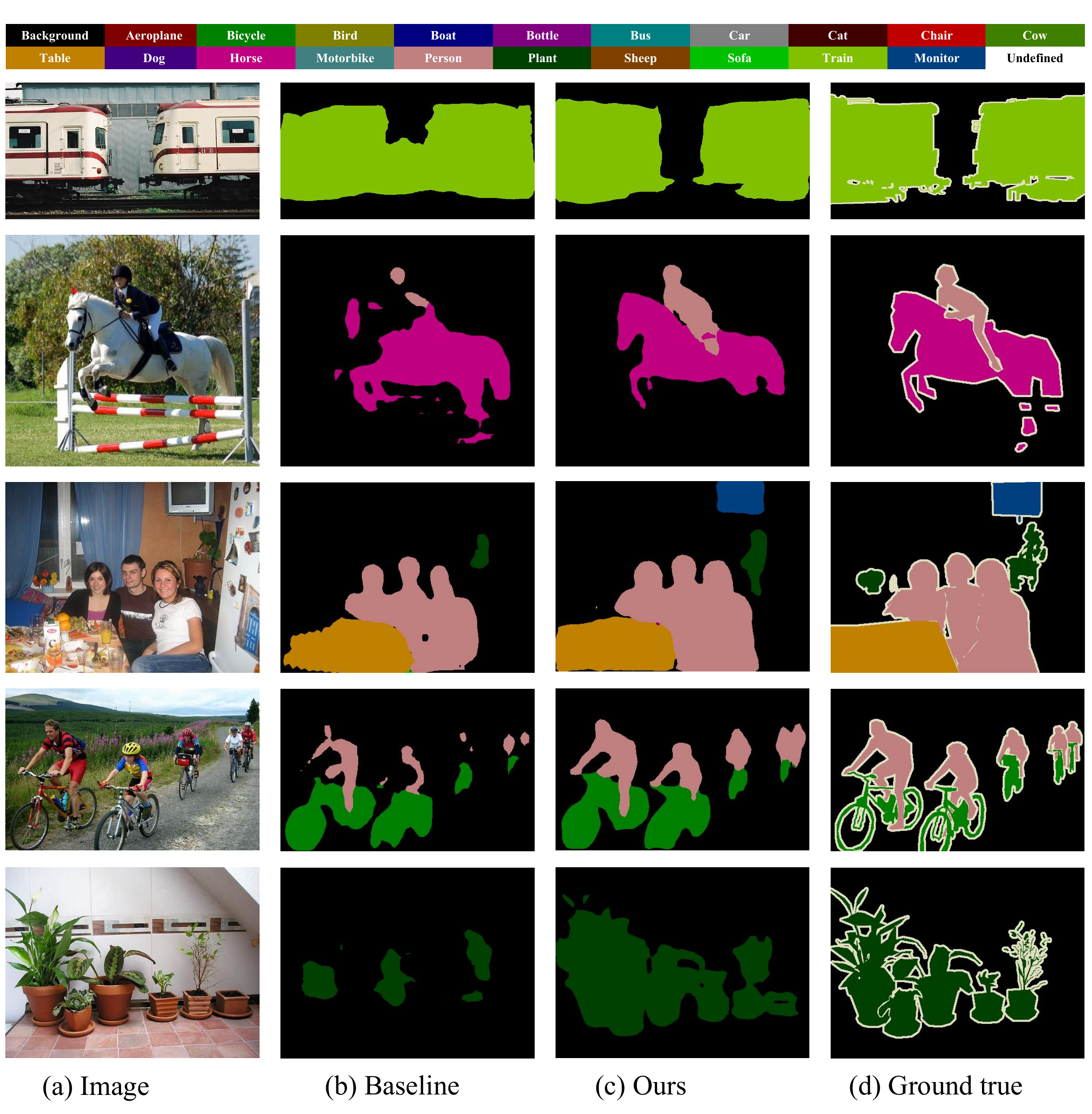}
			
		\end{center}
		
		\caption{Qualitative results of our semi-supervised semantic segmentation. The models are trained using 1/20 pixel-level labeled data from the training set of the PASCAL VOC 2012 dataset. All experiments are performed under DeepLabV2 and ResNet101 models. (a) and (d) are the images from the PASCAL VOC 2012 validation set and the corresponding ground-truth. (b) is the result of training with labeled data only. (c) is the result of training with our approaches.}
		\label{vocimage}
		
	\end{figure*}
	
	{\textbf{Effect of the size of memory bank. } 
		As shown in Table \ref{sizeofmemory}, when all the settings are the same, an extra-large size of memory bank size (such as 512) may benefit the performance, but not necessarily (such as 256 and 1024).  Note that $N$ = 0 means that HC and LC are removed from the framework.
		All the elements in the memory bank are used in the contrastive learning process, so the computation is proportional to the size of the memory bank. In the trade-off between performance and efficiency, we select 128 as the default value.
		
	}
	
	{\textbf{Effect of confidence threshold $\phi$. } 
		We investigate the influence of different $\phi$ used to distinguish between high-level contrastive learning and low-level contrastive learning, as shown in Table \ref{phi}. We find that $\phi=0.75$ results the best performance, and therefore this value is used in all other data partitions and architectures.
	}
	
	{\textbf{Effect of low-level contrastive learning. } 
		We investigate the influence of directional mask in low-level contrastive learning, as shown in Table \ref{directional}. When remove directional mask, it means that $m_{(A,B)}^{(i)}=(1-{m}_A^{(i)})$. Experiments prove the importance of direction.
	}
	
	\begin{figure*}[!htbp]
		
		\begin{center}
			
			\includegraphics[width=0.96\textwidth]{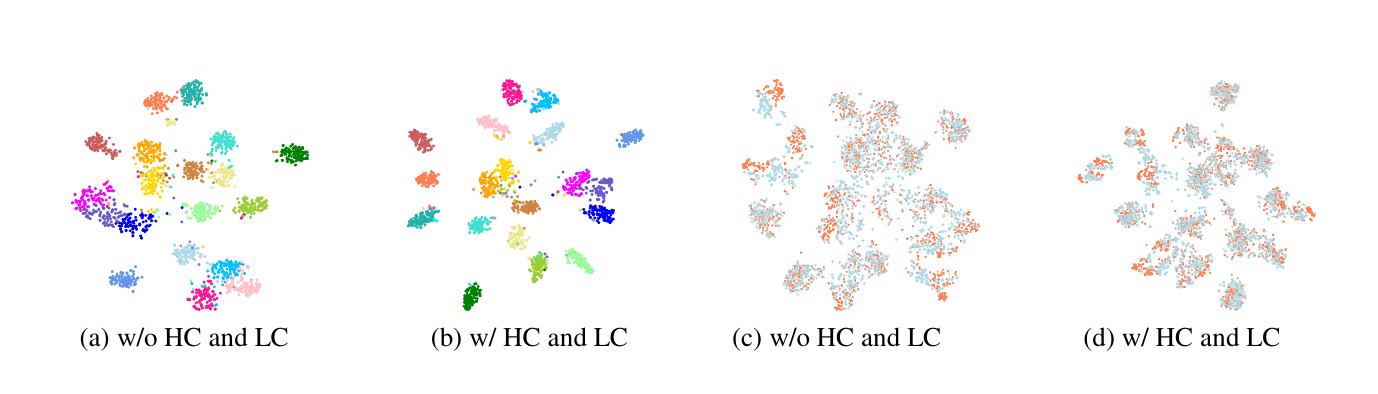}
			
		\end{center}
		
		\caption{Visualization of features, we use tSNE to map features extracted from input data to a 2D space for presentation purposes. (a) and (b) denote the distributions of different classes represented by different colors on the validation set of the Cityscapes. (c) and (d) represent the distributions of labeled (orange) and unlabeled (blue) data from the Cityscapes training set.
		}
		\label{class}
		
	\end{figure*}
	\begin{figure*}[!htbp]
		
		\begin{center}
			
			\includegraphics[width=0.96\textwidth]{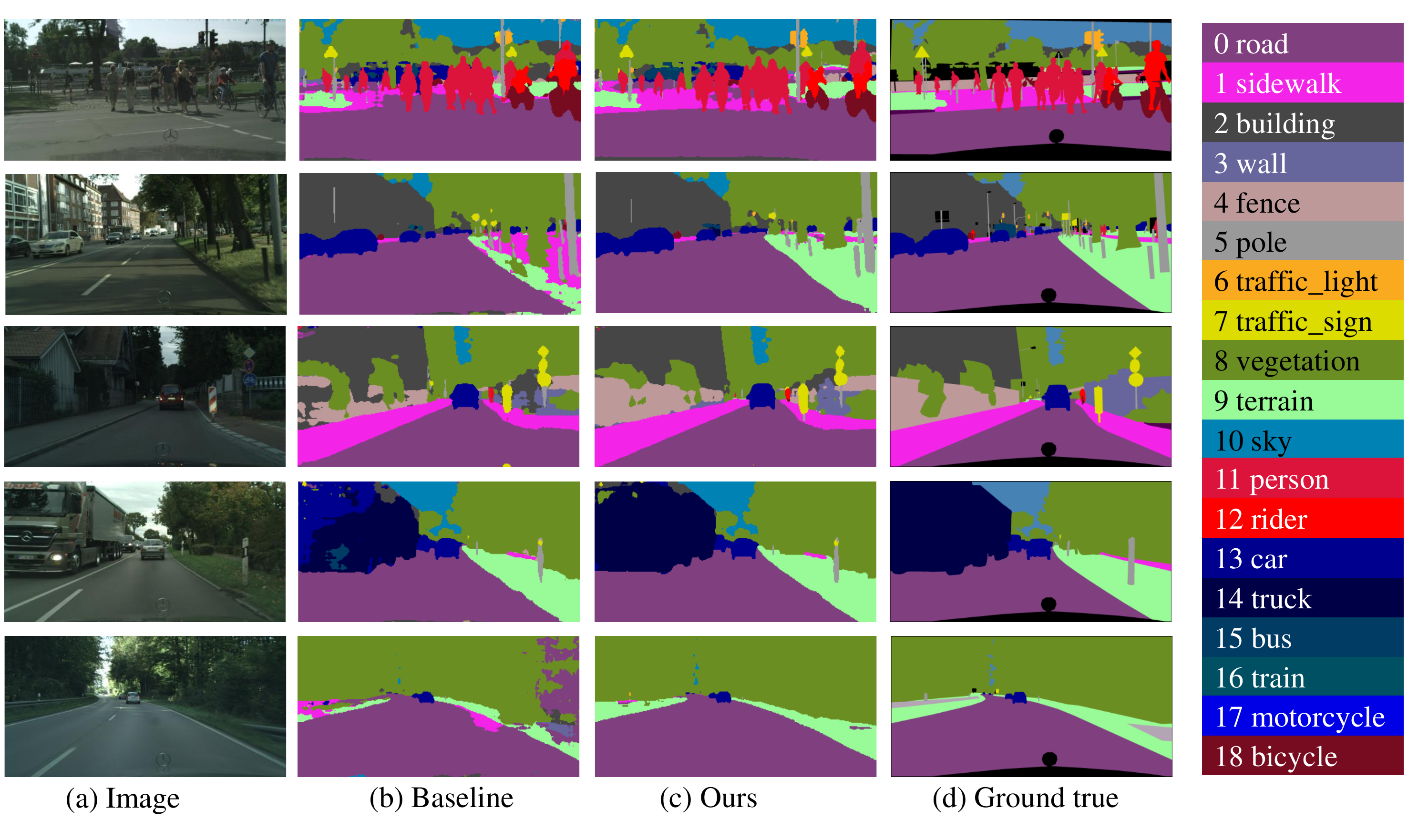}
			
		\end{center}
		
		\caption{Qualitative results of our semi-supervised semantic segmentation. The models are trained with 1/8 pixel-level labeled data from the training set of the Cityscapes dataset. All experiments are performed under DeepLabV2 and ResNet101 models. (a) and (d) are the images from the Cityscapes validation set and the corresponding ground-truth. (b) is the result of training with labeled data only. (c) is the result of training with CTT.}
		\label{image}
		
	\end{figure*}

	\textbf{Effectiveness of the cross-teacher module.}  To verify the effectiveness of the cross-teacher, we compare it to direct mutual teaching (which removes teacher networks and only two student networks teach each other), ensemble model (which combines the output of two separate pairs of student-teacher networks with ensemble learning), {self-training (which trains a teacher model with labeled data to
		produces pseudo labels for unlabeled data, and then retrains a new student model with labeled data and unlabeled data
		with pseudo-labels) \cite{xie2020self},} and mean teacher models \cite{tarvainen2017mean,olsson2021classmix}. In addition, we also
	compare with the setting of teaching each student with two
	teachers, which we refer to as dual teachers. As shown in Table \ref{cross-teacher ablation study}, our proposed cross-teacher outperforms direct mutual teaching, mean teacher, ensemble model, self-training and dual teachers, yielding increments of 3.34\%, 3.35\%, 2.12\%, 2.94\% and 1.27\% in mIoU, respectively. Mutual teaching enables adaptive correction of training errors through peer networks, but models are vulnerable to coupling with each other. The mean teacher can produce stable predictions but also suffer from model coupling problems \cite{ke2019dual}. Thus, our approach yields the best results because our strategy combines the advantages of co-teaching and mean teachers while reduces coupling between peer networks and avoids the error accumulation caused by the coupling of teacher and student networks.

	\begin{table}[!htbp]
		
		\setlength\tabcolsep{10pt}  
		\caption{Effectiveness of the cross-teacher module. We compare cross-teachers with direct mutual teaching, mean teachers, ensemble model, self-training, and dual teachers. Results are obtained on the Cityscapes dataset with $1/8$ labeled data, DeepLabV2-ResNet101.}
		\centering
		\label{cross-teacher ablation study}
		\scalebox{0.69}{
			\begin{tabular}{l|c}  
				\toprule   
				\textbf{Methods} & \textbf{mIoU\%} \\  
				\midrule
				\midrule   
				Baseline & 56.96 \\
				Direct mutual teaching & 60.35    \\  
				Mean teacher + ClassMix &  61.35   \\
				Ensemble model & 61.47 \\
				{ Self-training } & 60.75 \\   
				Dual teachers & 62.42 \\
				\midrule 
				\textbf{CTT  w/o $\mathcal{L}_{hc}$ and $\mathcal{L}_{lc}$} \,\,\,\,\,\,\,\,\,\,\,\,\,\,\,\,\,\,\,\,\,\,\,\,\,\,\,\,\,\,\,\,\,\,\,\,\,\,\,\,\,\,\,\,\,& \textbf{63.69}     \\
				\bottomrule  
			\end{tabular}
		}
	\end{table}

	\begin{table}[h]
		
		\caption{Effect of using n-pairs student-teacher network, where n represents the number of teacher-student networks in the training process. Results are obtained on the Cityscapes dataset with $1/8$ labeled data, DeepLabV2-ResNet101.}
		\centering
		\label{n-pairs}
		\scalebox{0.69}{
			\begin{tabular}{l|cccc}  
				
				\toprule   
				\textbf{n-pairs student-teacher network} & 1&2&3 \\  
				\midrule
				mIou\% & 60.34&63.69&\textbf{63.94} \\  
				\bottomrule  
			\end{tabular}
		}
	\end{table}

	\begin{table}[h]
		
		\setlength\tabcolsep{5pt} 
		\caption{Effect of memory bank size in our complementary contrastive learning (features per-class), $N$. Results are obtained on the Cityscapes dataset with $1/8$ labeled data, DeepLabV2-ResNet101.}
		\centering
		\label{sizeofmemory}
		\scalebox{0.69}{
			\begin{tabular}{l|cccccc}  
				
				\toprule   
				\textbf{Size ($N$)} &0& 64&128&256&512&1024 \\  
				\midrule
				mIou\% &56.96& 60.91&62.30&61.83&62.36&60.85 \\  
				\bottomrule  
			\end{tabular}
		}
	\end{table}

	\begin{table}[h]
		
		\setlength\tabcolsep{9pt} 
		\caption{Effect of confidence threshold $\phi$ in our complementary contrastive learning. Results are obtained on the Cityscapes dataset with $1/8$ labeled data, DeepLabV2-ResNet101.}
		\centering
		\label{phi}
		\scalebox{0.69}{
			\begin{tabular}{l|cccccc}  
				
				\toprule   
				$\phi$ & 0 & 0.75& 0.85 & 0.95 & 1 \\  
				\midrule
				mIou\% & 60.79 & \textbf{62.30} & 62.24 & 62.16 & 61.31\\  
				\bottomrule  
			\end{tabular}
		}
	\end{table}

	\textbf{Effectiveness of the complementary contrastive learning.} In this part, we explore the role of complementary contrastive learning for our semi-supervised learning framework. The purpose of this complementary contrastive loss is to help pixel features from the same class closer and features from different classes further. In Fig. \ref{class}(c) and Fig. \ref{class}(d), we can observe that our model increases the inter-class gaps and reduces the intra-class distances in comparison to without using the contrastive learning model. Furthermore, it prompts knowledge transfer from labeled to unlabeled data, reducing distribution gap between labeled and unlabeled data, as shown in Fig. \ref{class}(a) and Fig. \ref{class}(b).
	
	\begin{table}[h]
		
		\setlength\tabcolsep{6pt} 
		\caption{Effect of low-level contrastive learning in our complementary contrastive learning. Results are obtained on the Cityscapes dataset with $1/8$ labeled data, DeepLabV2-ResNet101.}
		\centering
		\label{directional}
		\scalebox{0.69}{
			\begin{tabular}{l|cc}  
			
				\toprule   
				Strategy & w/o directional mask & w/ directional mask \\  
				\midrule
				mIou\% & 61.69 & \textbf{62.30}\\  
				\bottomrule  
			\end{tabular}
		}
	\end{table}

	\subsection{Visualization Results}
	We show the qualitative results of our method on the Cityscapes dataset  in Figure \ref{image} and PASCAL VOC 2012 dataset in \ref{vocimage}. Obviously, our method has a remarkable improvement over training on labeled data only. Our method effectively improves the effectiveness of semi-supervised training.

	\section{Conclusion}
	In this paper, we proposed a new framework for semisupervised semantic segmentation. The cross-teacher training
	framework reduces coupling between networks, that can reduce error accumulation during training comparing with traditional approaches. We also propose two complementary
	contrastive learning modules to facilitate the learning process and address the limitation of cross-entropy. One module
	helps create better class separation across the feature space
	and transfer knowledge from labeled data to unlabeled ones.
	The other encourages networks to learn high-quality features
	from peer networks. Our approach can be readily adopted by
	a variety of semantic segmentation networks. Experimental
	results demonstrate that our approach achieves superior performance than state-of-the-art methods.
	
	\section{Acknowledgment}
	This work was supported by the Natural Science Foundation of Zhejiang Province (NO. LGG20F020011), Ningbo Public Welfare Technology Plan (2021S024), Ningbo Science and Technology Innovation Project (No. 2021Z126),  the National Natural Science Foundation of China (61901237, 62171244), Alibaba Innovative Research Program, and Open Fund by Ningbo Institute of Materials Technology \& Engineering, the Chiense Academy of Sciences.

	\bibliographystyle{elsarticle-num-names}

	\bibliography{cas-refs}

\end{document}